\documentclass[journal]{IEEEtran}
\pdfoutput=1
\usepackage{amsmath,amssymb,amsfonts}
\usepackage{graphicx}
\usepackage{mathptmx}
\usepackage[font={small}, labelfont=bf]{caption}
\usepackage{textcomp}
\usepackage{xcolor}
\usepackage{bm}
\usepackage{amsmath}
\usepackage{amssymb}
\usepackage{amsthm}
\usepackage{mathrsfs}
\usepackage{enumerate}
\usepackage{multirow}
\usepackage{color}
\usepackage{threeparttable}
\usepackage{lipsum} 

\usepackage[sort,square, comma, sort&compress, numbers]{natbib}
\usepackage[pagebackref=false,breaklinks=true,colorlinks,bookmarks=false]{hyperref}

\usepackage{times}
\usepackage{url}
\usepackage{array}
\usepackage{verbatim}
\usepackage{bbding}
\usepackage{algpseudocode}
\usepackage{lettrine}
\usepackage{stfloats}  
\usepackage{authblk}

\def\BibTeX{{\rm B\kern-.05em{\sc i\kern-.025em b}\kern-.08em
    T\kern-.1667em\lower.7ex\hbox{E}\kern-.125emX}}
\begin{document}

\title{A Multi-scale Residual Transformer for   VLF Lightning Transients Classification \\}

\author[1]{Jinghao Sun}
\author[*2]{Tingting Ji\thanks{*Corresponding author  
Email address: tinaji$\_$ouc@163.com}}
\author[3]{Guoyu Wang}
\author[4]{Rui Wang}
\affil[1]{Faculty of Information Science and Engineering, Ocean University of China, Qingdao, China}
\affil[2]{Teaching Center of Fundamental Courses, Ocean University of China, Qingdao, China}
\affil[3]{Faculty of Information Science and Engineering, Ocean University of China, Qingdao, China}
\affil[4]{Physics and Electronic Information, Yantai University, Yantai, China}

\maketitle

\begin{abstract}

The utilization of Very Low Frequency (VLF) electromagnetic signals in navigation systems is widespread. However, the non-stationary behavior of lightning signals can affect VLF electromagnetic signal transmission.  Accurately classifying lightning signals is important for reducing interference and noise in VLF, thereby improving the reliability and overall performance of navigation systems. In recent years, the evolution of deep learning, specifically Convolutional Neural Network (CNNs), has sparked a transformation in lightning classification, surpassing traditional statistical methodologies. Existing CNN models have limitations as they overlook the diverse attributes of lightning signals across different scales and neglect the significance of temporal sequencing in sequential signals. This study introduces an innovative multi-scale residual transform (MRTransformer) that not only has the ability to discern intricate fine-grained patterns while also weighing the significance of different aspects within the input lightning signal sequence. This model performs the attributes of the lightning signal across different scales and the level of accuracy reached 90$\%$ in the classification. In future work, this model has the potential applied to a comprehensive understanding of the localization and waveform characteristics of lightning signals.
\end{abstract}


\section{Introduction}
\IEEEPARstart{V}{ery} low frequency (VLF) ~\cite{arshad2019lightning} ~\cite{konan2020machine} refers to long-wave communication using radio waves ranging between 3kHz and 30kHz (wavelengths from 100km to 10km). In comparison to communication operating within the microwave frequency range, terrestrial long-wave communication within the VLF band presents distinct advantages. These include channel stability, extended propagation distances, capacity for water penetration, and the ability to navigate intricate terrain. Such advantages have found wide-ranging applications across diverse domains ~\cite{golkowski2018ionospheric, fu2022bag, weng2023cross, hosseini2019remote}, spanning submarine communication to satellite navigation systems.

Leveraging VLF electromagnetic signals offers a notably stable approach to remote sensing. However, as these signals propagate through the 'earth-ionosphere wave-guide,' atmospheric noise interference arising from various activities emerges as a primary challenge for VLF signals ~\cite{liu2019large, harid2021automated, rapoport2020model, hu2023bag, hu2023robust}. Among these sources, lightning discharges proximate to the signal receiver emerge as the most substantial interference source. Lightning impulses introduce non-stationary characteristics within VLF, consequently diminishing communication quality and impacting navigational positioning accuracy.
The classification of lightning events allows for capturing the fundamental waveform features of nearby lightning impulses, serving as the focal point in research endeavors pertaining to lightning signal detection, localization, and the enhancement of communication quality within VLF sensing.
\begin{figure}[htb]
    \centering
    \includegraphics[width=1\linewidth]{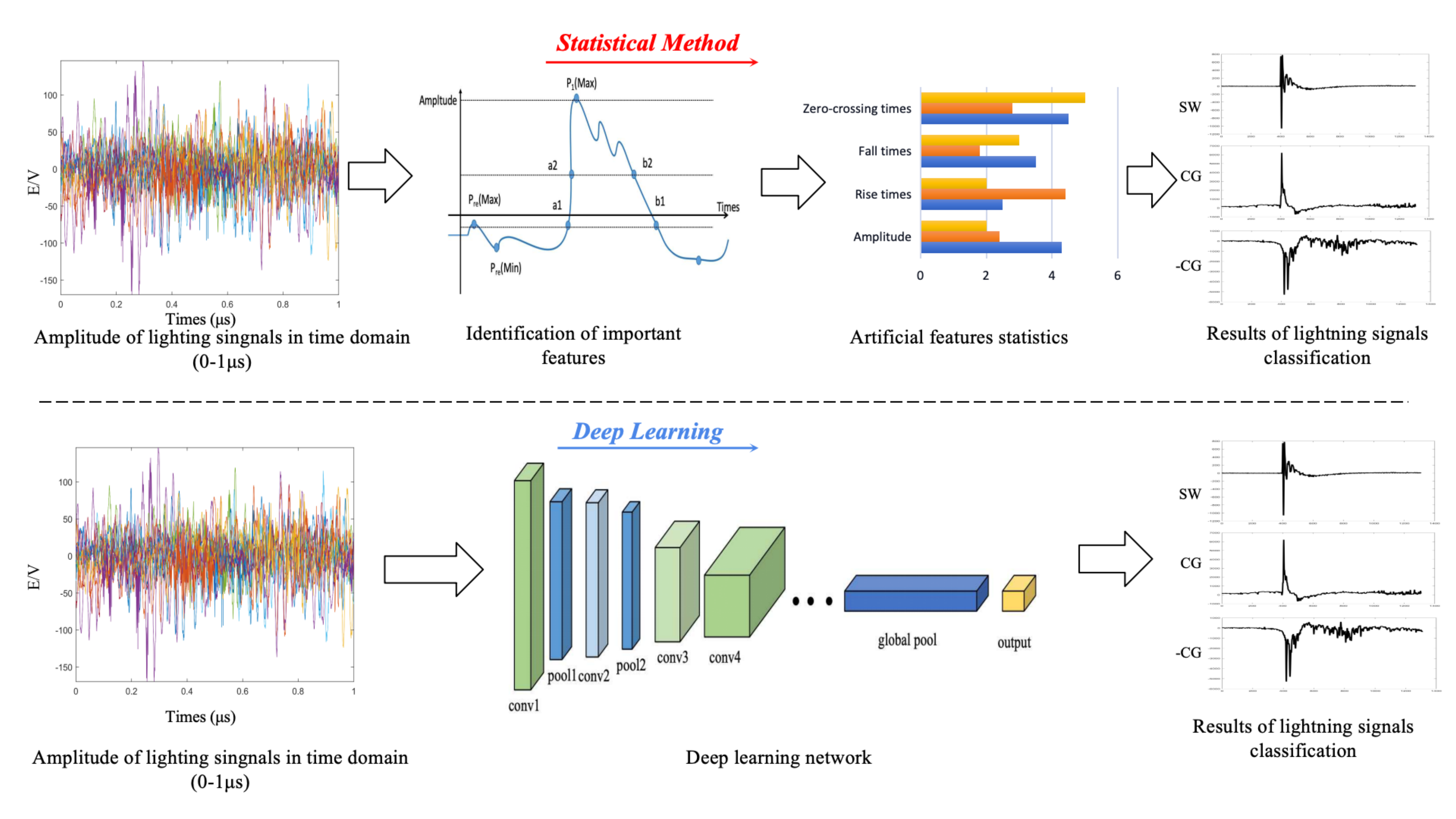}
    \captionsetup{justification=centering}
    \caption{The comparison schema for statistical method and deep learning}
\label{fig:schema}
\end{figure}

VLF sensing and the Time of Arrival (TOA) method have played a crucial role in lightning location systems. Traditionally, statistical methods have been used to determine important features of VLF waveforms, such as amplitude, rise and fall times, zero-cross times, etc ~\cite{cooray2004fine}~\cite{smith2002alamos}. However, in practical applications, there is a wide range of lightning types and a large amount of data to handle. The multi-parameter method faces challenges in extracting characteristic parameters from low-amplitude VLF signals, often resulting in the neglect of these smaller signals and reducing detection efficiency ~\cite{kohlmann2017evaluation}~\cite{nag2014recent}. Moreover, the characteristic parameters used in the multi-parameter method may vary across regions with different meteorological conditions ~\cite{said2010long, shen2022hsgnet, cooray2009propagation, wooi2015comparative}. Therefore, the multi-parameter method is not an ideal approach for lightning signal classification. As shown in  Fig.\ref{fig:schema}, deep learning networks are introduced to improve the classification efficiency of lightning VLF/LF signals, particularly in managing intricate datasets and elevating classification precision. The deep learning techniques for the classification of lightning events remain a relatively unexplored domain. Previous studies have much more effect on classifying lightning signals by using CNN networks and that with different structures.~\cite{wang2020classification}~\cite{peng2022lightning}~\cite{peng2019convolutional}. ~\cite{xiao2023towards} gave an explanation for CNN networks. While these studies have ignored the impact of long-range dependencies in lighting sequences, understanding the relationships between distant parts of the signal is an essential part of classifying the lightning signals.
\begin{figure*}[htb]
    \centering
    \includegraphics[width=1\linewidth]{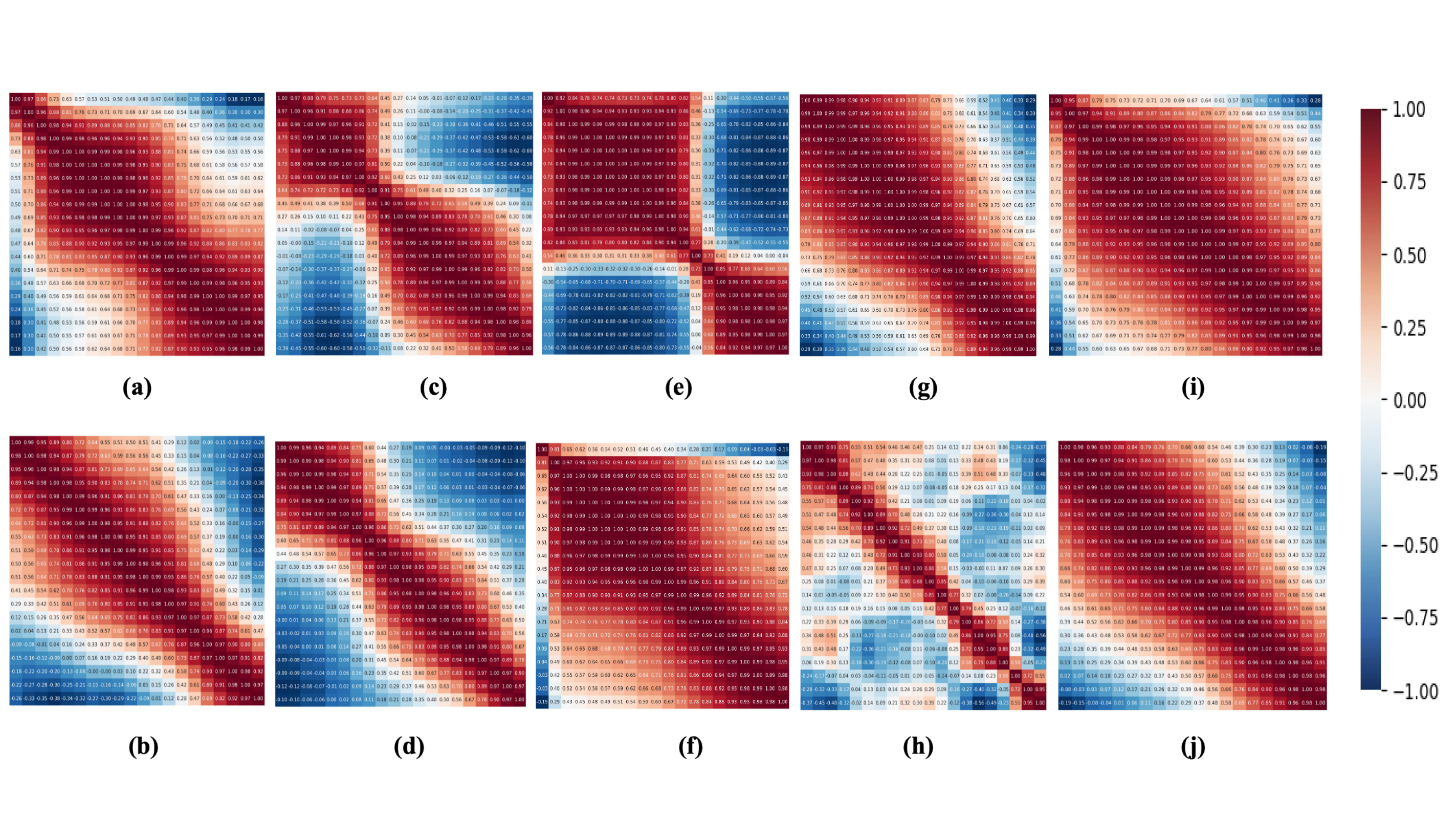}
    \caption{The correlation heatmaps of lighting signals:(a)negative cloud-to-ground flash(-CG);(b)positive cloud-to-ground flash(+CG);(c)negative pre-breakdown(-PBP);(d)positive PBP(+PBP);(e)negative narrow bipolar envet(-NBE);(f)positive NBE(+NBE);(g)NBE;(h)multi-pulse cloud flash(MP);(i)cloud ground flash with ionosphere reflected signals(CG-IR);(j)skywave(SW).}
    \label{fig:correlation heatmaps}
\end{figure*}
There have also been previous studies ~\cite{kandalgaonkar2003diurnal} ~\cite{geng2019lightnet} that have explored the use of time-series-related methods for predicting lightning signals. It has been observed, as shown in Fig.\ref{fig:correlation heatmaps}, that a majority of lightning signals display a strong correlation in the time series. Therefore, considering the correlation in time series is a key factor in the classification of lightning signals.

This article proposes a \textit{multi-scale residual transformer} (MSRT) model to classify VLF lighting signals. The architecture comprises a multi-scale model with a connected residual and transformer model. To train this model, an open-source dataset including ten types of lighting signals was established for model training.  Subsequently, The classification accuracy of the proposed model was compared with state-of-the-art methods using F1 score as evaluation criteria. To ensure a comprehensive comparison, we assessed the classification accuracy of each type by employing the Receiver Operating Characteristic curve analysis for two selected categories of lighting signals. In the ablation study, we utilized 5-fold cross-validation to assess the performance of our model. Additionally, we compared our model's performance to that of a transformer-based model using accuracy and F1 score as metrics. To provide further insight into our model's performance, we visualized the lighting features at each multi-scale layer using histograms. Finally, the generalization of the proposed model was tested on a dataset collected from Xinjiang.

In essence, this paper presents three significant contributions. Firstly, it introduces a multi-scale residual module that effectively captures information across multiple scales. This module enhances the model's capability to discern local nuances within lightning signals at varying scales and skillfully integrates them to capture the overarching characteristics of lightning signals. Secondly, the paper combines the multi-scale residual module with a transformer model, marking the first instance of integration in VLF lightning signal classification. Furthermore, this approach takes into account the long-range dependencies present at each point of the lightning signal sequence.

\section{The Related Work}

\subsection{The Related Lighting Signal Classification Work}

\begin{figure}[htb]
    \centering
    \includegraphics[width=1\linewidth]{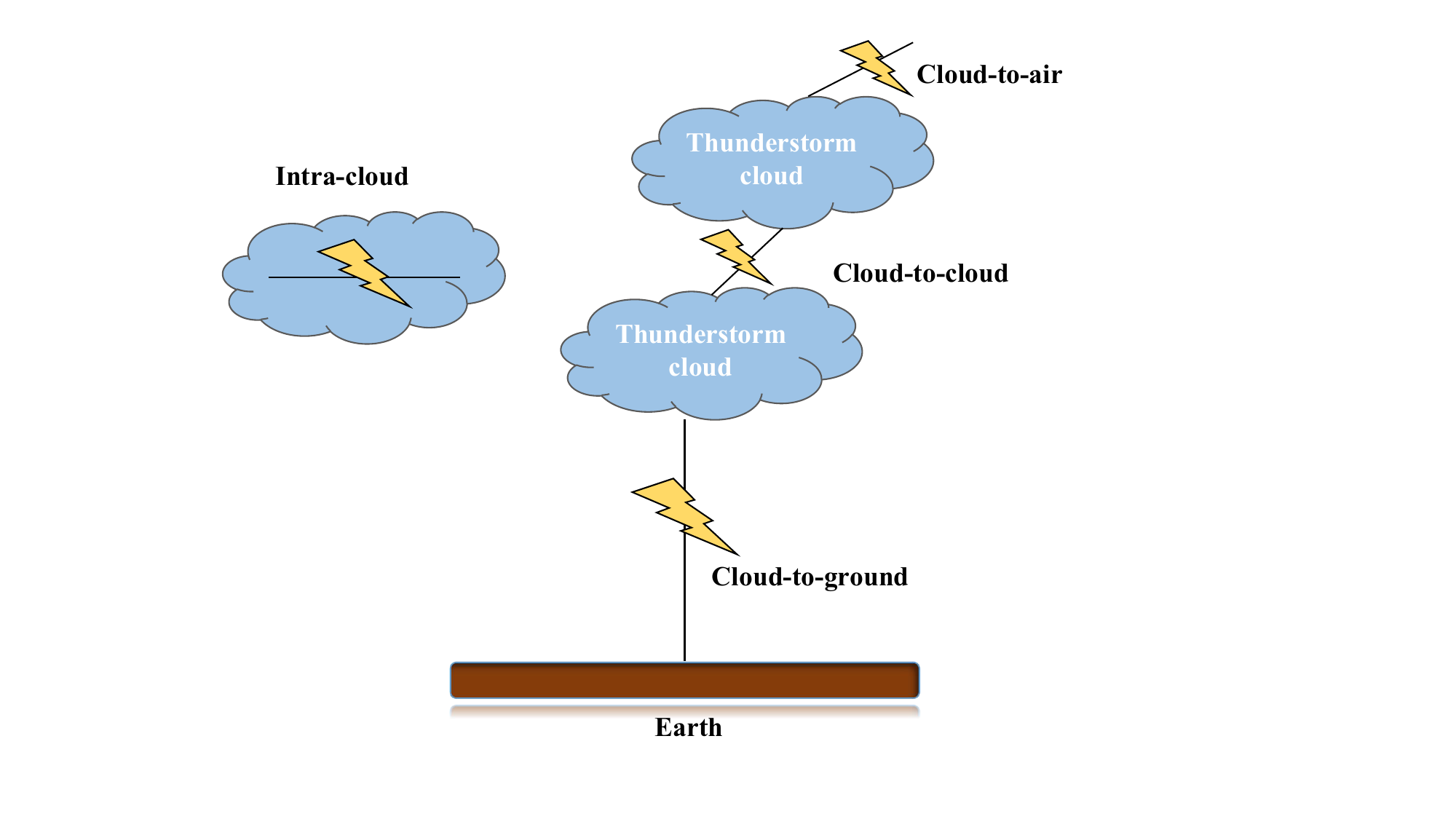}
    \caption{The multiple types of lighting such as ground wave, sky wave,and wave conduction.}
    \label{fig:typical lighting discharge}
\end{figure}

Lightning, a dynamic atmospheric discharge,~\cite{maslej2021automatic} ~\cite{said2010long}generates electromagnetic radiation detectable in the VLF. Harnessing this phenomenon for lightning detection and location is crucial for minimizing risks associated with lightning strikes. Based on the location where lightning is produced, lightning can be divided into two types: cloud-to-ground lightning (CG) and cloud lightning (CC).~\cite{zhu2021machine}~\cite{davis2008enhancement} CG refers to lightning that occurs between thunderclouds and the earth or objects on the earth. CC refers to the entire process of lightning discharge without contacting objects on the ground. Such as the Fig.\ref{fig:typical lighting discharge} below, CC can be further divided into inter-cloud lightning, intra-cloud lightning, and cloud-air lightning. These emissions provide a unique signature that can be exploited for lightning detection and classification.

Classifying lightning events accurately is a pivotal challenge in lightning detection. The study ~\cite{hepburn1957wave} on waveform correlation of lightning pulses in VLF found that lightning strikes in this frequency range typically exhibit oscillating short pulses with multiple peaks and valleys, lasting from 1 to 10 milliseconds. Malan ~\cite{malan1963physics} used optical instruments and photography to study ground lightning and discovered that a single ground lightning is composed of multiple pulses, each lasting from several to tens of milliseconds, with a total duration of about 0.2 seconds. Alpert ~\cite{alpert1967propagation} and Challinor~\cite{challinor1967phase}  also recorded transient photography of lightning pulses in the VLF range.

Traditional methods employ multi-parameter criteria~\cite{wang2020classification, cai2013ground}, but these often struggle with low-amplitude signals and variations across meteorological conditions. The need for precise classification is underscored by the potential consequences of misclassification, making it imperative to discern between CGs, ICs, and other events like narrow bipolar events (NBEs) ~\cite{bitzer2013characterization, wang2016beijing, wu2018locating}.

In recent years, machine learning methods, such as Support Vector Machines (SVM) and Convolutional Neural Networks (CNNs)~\cite{gao2020classification, shen2023triplet, peng2022lightning, wang2020classification}, especially ~\cite{xiao2023towards} proposed an interpretable CNN have gained prominence in lightning signal classification. However, the existing CNN models have not fully considered the different characteristics of lightning signals at different scales and the temporal correlation of lightning signals.

The field of lightning signal processing is rapidly evolving, and the accurate classification of lightning events remains a focal point. Advancements in deep learning techniques hold promise, but the challenges of data imbalance and regional variability must be addressed. Developing multi-scale learning models can bridge the gap between features and long-range dependencies in lightning sequences to accommodate the diversity of lightning environments, enable reliable lightning event classification, and ultimately improve the effectiveness of lightning detection systems.

\subsection{The Role of Deep Learning Classification Networks}

Deep learning classification networks \cite{shen2022competitive, qiao2022novel, shen2023pbsl} have a broad and impactful role in various application areas beyond lightning signal processing. They excel in tasks like image classification~\cite{krizhevsky2012imagenet}, object recognition \cite{li2022enhancing, xu2021dual}, and medical diagnosis by recognizing patterns in data that might be missed by humans. These networks also play a crucial role in the natural language processing ~\cite{devlin2018bert}, object detection \cite{qiao2022novel}, and re-identification \cite{wu2022sample, shen2023triplet}. For example, Graph interactive Transformer (GiT) \cite{shen2023git}  proposes a structure where graphs and transformers interact constantly, enabling close collaboration between global and local features for vehicle re-identification. Hybrid Pyramidal Graph Network (HPGN) \cite{shen2021exploring}  proposes a novel pyramid graph network targeting features, which is closely connected behind the backbone network to explore multi-scale spatial structural features.  Hierarchical Similarity Graph Module (HSGM) \cite{shen2022hsgm} proposes a hierarchical similarity graph module to relieve the conflict of backbone networks and mine the discriminative features.

Additionally, deep learning networks contribute to in drug discovery, environmental monitoring, and quality control in manufacturing. Overall, deep learning classification networks have transformative potential across a wide range of fields and industries. Their ability to extract meaningful patterns from complex data has led to advancements in automation, accuracy, and efficiency in various applications.

\section{Methods}

\begin{figure*}[htb]
    \centering
    \includegraphics[width=1\linewidth]{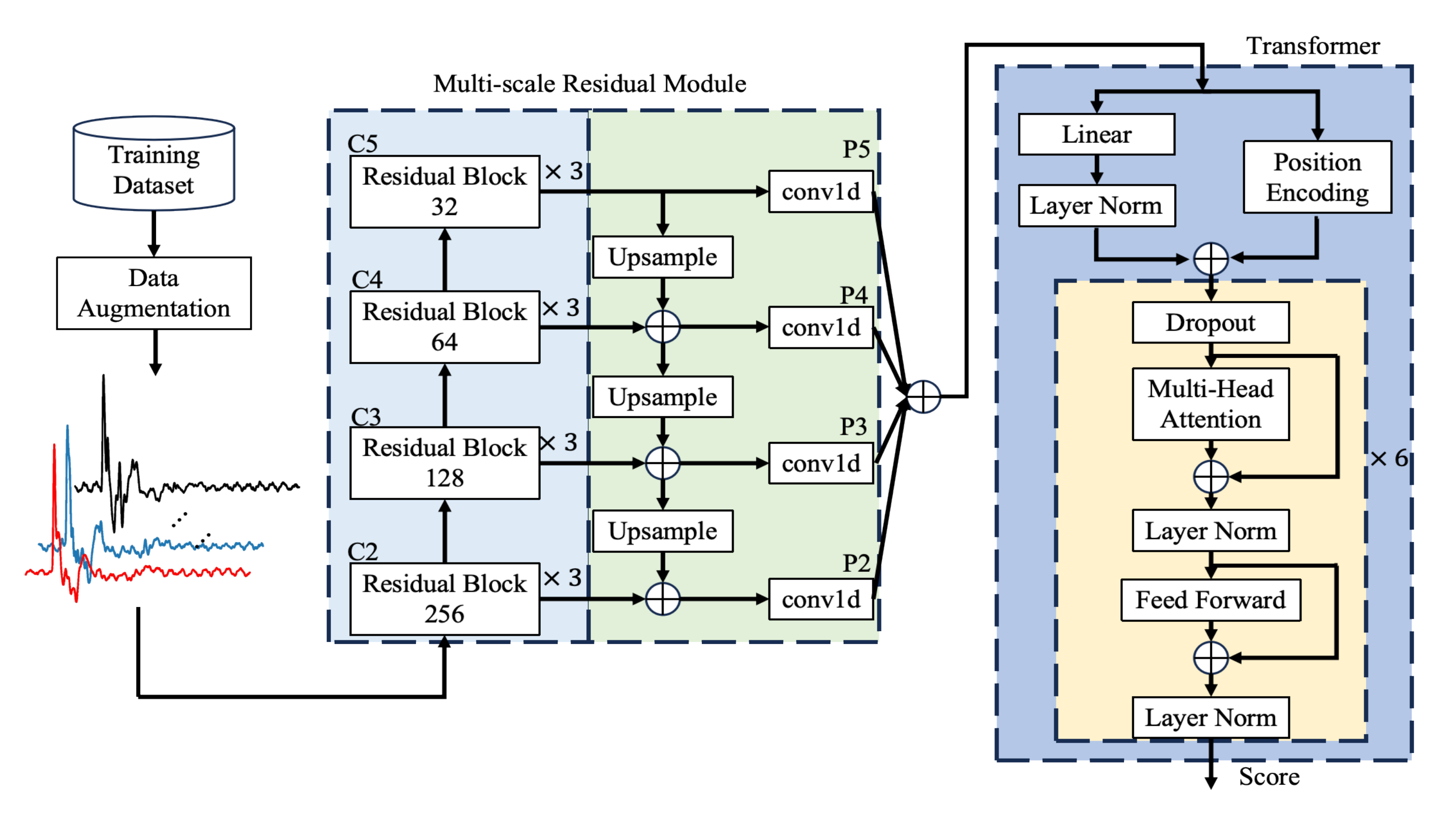}
    \caption{The proposed model structure is as follows: Steps 1-3 describe the input waveform to the multi-scale module, which is then connected to the residual model through the transformer model. Finally, the model outputs the classification.}
    \label{fig:net_structure}
\end{figure*}

MSRT is a CNN-Transformer hybrid signal classification model, which can simultaneously learn global and local feature representations and capture long-range information at ease. Concretely, MSRT mainly consists of two parts: the Multi-scale Residual Module is designed to extract local features of the input signal. The transformer Module captures global information. The overall framework is depicted in Fig.\ref{fig:net_structure}.

\subsection{The Classical Classification Networks}

CNN networks have a wide range of applications in classification, but they also have certain limitations. Firstly, CNN networks face problems such as gradient vanishing, gradient explosion, or network degradation, leading to insufficient model generalization ability. Therefore, He et al.~\cite{he2016deep} proposed the ResNet structure based on residual blocks and achieved highly competitive classification results on multiple datasets. 

Secondly, each convolutional layer has the same size as the convolutional kernel, which makes it difficult to handle the possible multi-scale features in the lightning waveform.

Finally,  CNN networks not consider the temporal correlation of time sequence signals. In 2017, the Google research team proposed the Transformers architecture based on the self-attention mechanism~\cite{vaswani2017attention}, which showed great advantages in sequence modeling. Then, the Shifted Window Multi-Head Attention (SW-MSA) is used to achieve information fusion between non-overlapping windows. 

Hence, in this paper, to address the limitation of CNN networks, there is a multi-scale transformer network proposed.

\subsection{The Multi-scale Residual Transformer Structure}
As shown in  Fig.\ref{fig:net_structure}, the model consists of several components, each playing a crucial role in ensuring its effectiveness. One of these components is the \textit{multi-scale residual} (MSR) model, which is designed with residual parts to capture information at multiple scales. The MSR network architecture uses a Feature Pyramid Network (FPN)~\cite{lin2017feature} backbone on top of a feedforward ResNet architecture to generate a rich feature pyramid. This model adeptly tackles this challenge by intelligently capturing information at multiple scales, ensuring that no crucial detail goes unnoticed. By integrating these diverse features, the model becomes capable of making decisions that are not only well-informed but also consider the holistic character of the signals under analysis. This feature integration mechanism is particularly beneficial in scenarios where signals manifest intricate patterns across different scales, enhancing the overall classification accuracy.

After the MSR, the transformer encoder is employed to handle long-range dependencies of each point of lighting signal sequences. The Transformer Encoder, with its inherent self-attention mechanisms, offers a solution to the complexities posed by long-range dependencies within lighting signal sequences. These self-attention mechanisms enable the model to weigh different parts of the input sequence, facilitating better contextual understanding and capturing intricate patterns within the data. By utilizing attention mechanisms, the transformer encoder can identify subtle relationships within the signal data, leading to accurate classification results.

The transformer encoder module consists of two parts: 1) a multi-head self-attention mechanism after a layer normalization, and 2) a multi-layer perception after multi-head self-attention and a layer normalization. In addition, in order to capture rich feature information, a residual connection is employed to connect these parts. The self-attention can be calculated as:
\begin{equation}
\label{eq:self-attention}
Attention(Q,K,V) = SoftMax(\frac{{QK}^{T}}{\sqrt{d}})V
\end{equation}

As shown in Fig.\ref{fig:net_structure}, the multi-head attention comprises 6 self-attention blocks. After multi-head attention, the Feed-Forward Network (FFN) presents the capability to leverage local context. Following the intricate processing of features, the architecture employs a  layer normalization step. This step ensures that the intermediate outputs remain consistent and well-behaved throughout the network. Layer normalization not only contributes to the stability of the training process but also enhances the overall speed of convergence. 

Finally, the model employs a Linear fully connected layer as the final component to generate the classification output. This layer plays a crucial role in bridging the processed features and the desired classification outcome.

\subsection{The Multi-scale Residual Module}

As shown in Fig.\ref{fig:net_structure}, the MSR is an essential component of the proposed model architecture, designed to capture information at multiple scales. We adopt the Feature Pyramid Network (FPN) as the backbone network for the MSR, build FPN on top of the Residual Network. The MSR module constructs a rich and multi-scale feature pyramid from a 1D signal by incorporating a top-down pathway and lateral connections. As the Equation \ref{eq:residual block} description as:
\begin{equation}
\label{eq:residual block}
{C}^{i+1}=\mathcal{F}(x,\left \{ {w}_{i}\right \})
\end{equation}
Where ${C}^{i+1}$ is defined as the output of the residual block, $x$ is defined as the input lighting signal. $\mathcal{F}(x,\left \{ {w}_{i}\right \})$ represents the residual mapping to be learned. The purpose of these residual layers is to downsample the input data and prevent gradients from disappearing.

The MSR extracts the features ranging from P2 to P5 of the feature pyramid. Description as the Equation \ref{eq:Upsampling}:
\begin{equation}
\label{eq:Upsampling}
{P}^{i} = conv1D({w}_{i},\left [Upsampling(C^{i+1})+C^{i}\right ]) + {b}^{i}
\end{equation}

Where $P^{i}$ is the output of the $i$ layer, $C^{i+1}$ through $Upsampling$ layer to extend the output length. The upsampling operation aligns features from different scales, ensuring that information from multiple scales is combined effectively. Then multiply with $C^{i}$, which is through convolution. $w_{i}$ is the multi-size convolution kernels at layer $i$.  ${b}^{i}$ is defined as the bias layer $i$. The output is the result of integrating the multi-scale features of each layer. This pathway allows the model to improve its understanding of the signal by first processing higher-level information, which then guides the processing of lower-level details. This iterative refinement mechanism enables the model to progressively fine-tune its understanding, leading to more accurate feature extraction.

\section{Dataset Description}

\subsection{The Experimental Dataset}
In the experiment, we utilized the dataset from reference\cite{wang2020classification}. The open-source lightning data used in this study was collected by the Institute of Electrical Engineering, Chinese Academy of Sciences, using the advanced direction-time lightning detection system (ADTD). This system encompasses 333 detection stations across China and Southeast Asia, equipped with a channel bandwidth of 3 kHz to 400 kHz. The system's GPS time synchronization accuracy is within 20 ns, ensuring precise event timestamps. For locations with waveform acquisition, trigger sampling is employed. This method achieves a sampling rate of 1 MSPS, capturing lightning characteristics with 1 ms precision. Additionally, a pre-trigger sampling window of 100 \(\mu s\) captures data leading up to the lightning event.
\begin{figure*}[htb]
    \centering
    \includegraphics[width=1\linewidth]{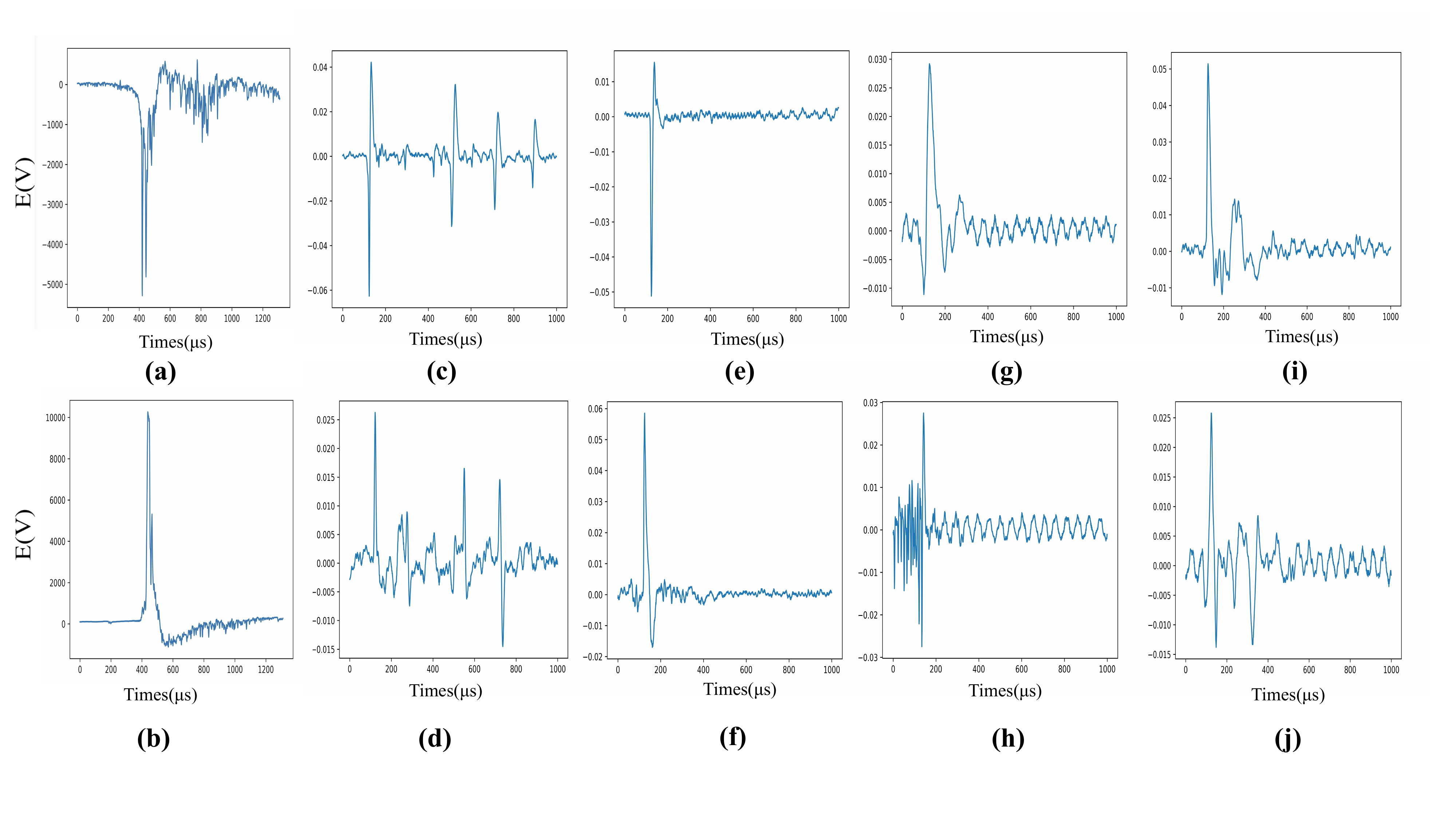}
    \caption{The 10 types of signals: (a)negative cloud-to-ground flash (-CG); (b)positive cloud-to-ground flash (+CG); (c)negative pre-breakdown (-PBP); (d)positive PBP(+PBP); (e)negative narrow bipolar envet(-NBE); (f)positive nbe(+NBE); (g)NBE; (h)multi-pulse cloud flash (MP); (i)cloud ground flash with ionosphere reflected signals (CG-IR); (j)skywave(SW).}
    \label{fig:typical VLF}
\end{figure*}
According to reference,these signals were further divided into 10 types: negative cloud-to-ground flash (-CG), positive cloud-to-ground flash (+CG), negative pre-breakdown (-PBP), positive PBP (+PBP), negative narrow bipolar envet (-NBE), positive NBE (+NBE), NBE, multi-pulse cloud flash (MP), cloud ground flash with ionosphere reflected signals (CG-IR) and skywave (SW). Fig. \ref{fig:typical VLF} shows the typical waveforms of these types. The total number of signal samples that have been used is around 12000. 

\subsection{Testing Dataset Description}
\begin{figure}[htb]
    \centering
    \includegraphics[width=1\linewidth]{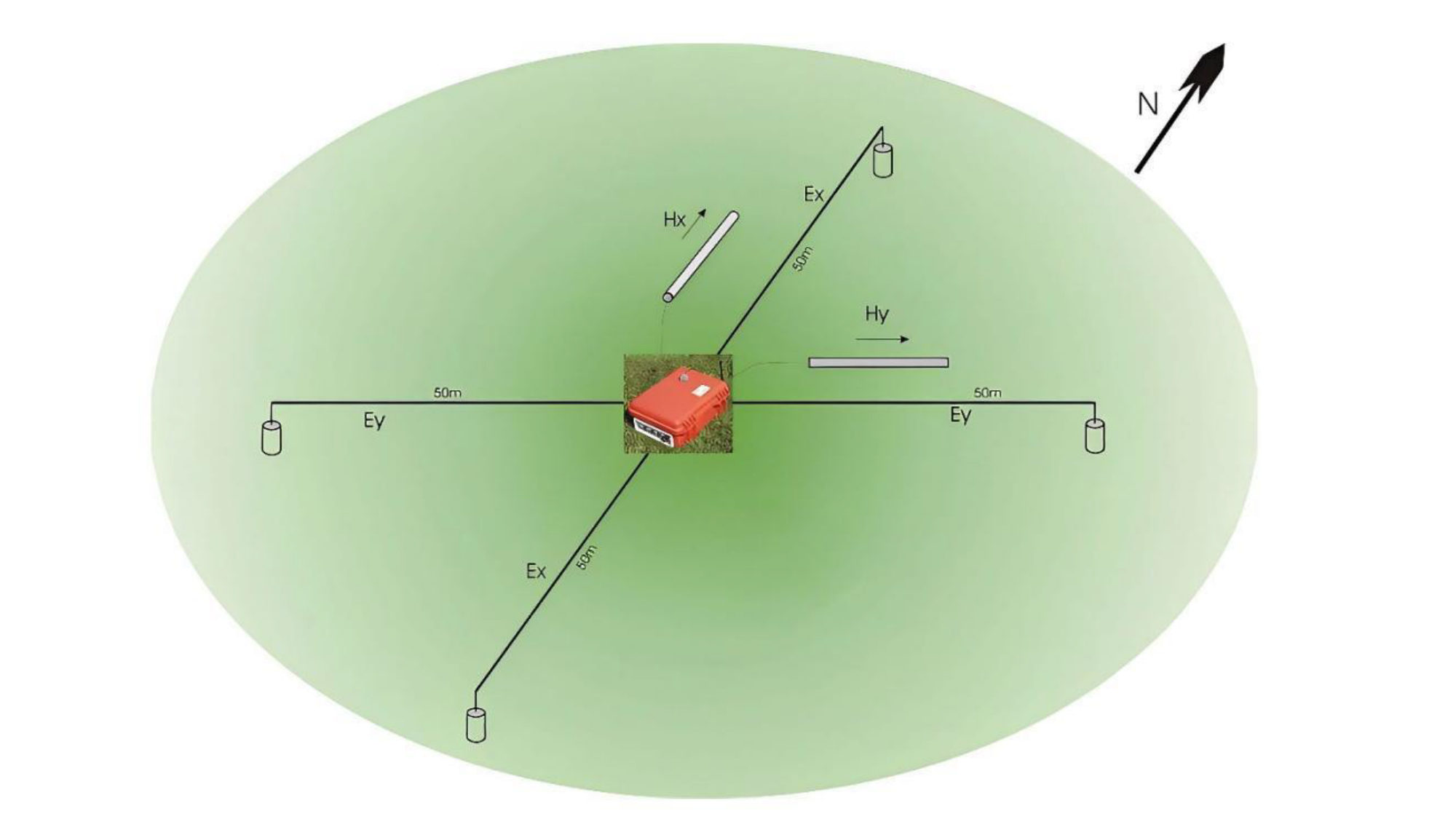}
    \captionsetup{justification=centering}
    \caption{The receiver for our implement dataset}
\label{fig:Receiver}
\end{figure}

\subsubsection{The Testing Dataset}

The Yili of Xinjiang was studied using the GMS-08e electromagnetic instrument, and actual data was obtained. The MFS-07e magnetic field sensor was utilized with a 50KHz bandwidth for every 1000 seconds. The VLF data collection system, shown in Fig. \ref{fig:Receiver} , featured the ADU-08e host in the middle, the Hx and Hy long bars as magnetic field sensors aligned in the north-south and east-west directions, and four small cylinders as non-polarized electrodes. The data acquisition equipment was set to a sampling rate of 131072Hz. In the implement experiment, the real data manually set a threshold of 500 millivolts was categorized as CG, CC, NBE, MP, or others in a total around of 900  data instances.

\subsubsection{Testing Dataset Preprocessing}

\begin{figure}[htb]
    \centering
    \includegraphics[width=1\linewidth]{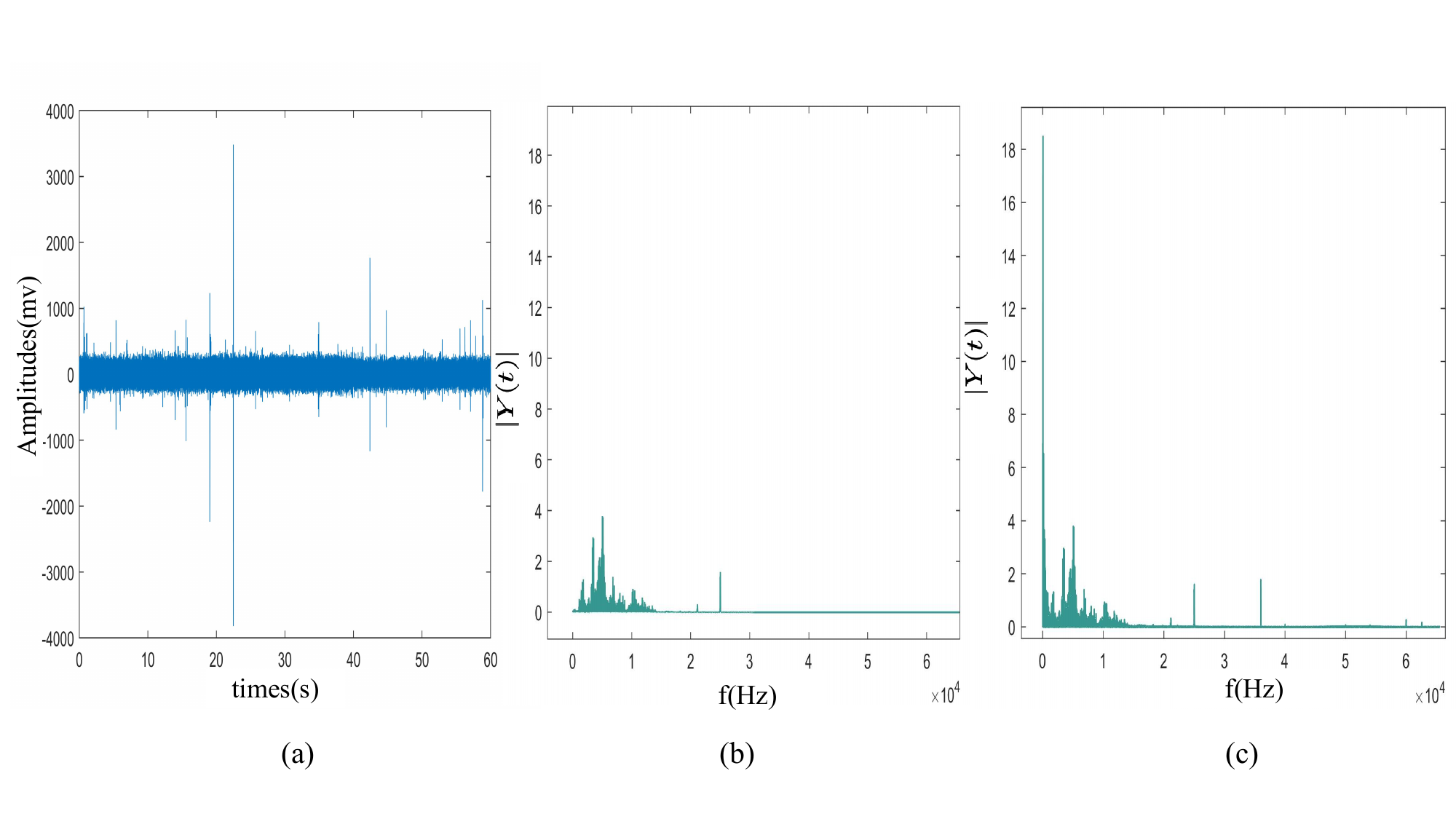}
    \caption{The results of preprocessing:(a) the original lighting signal;(b)the frequency domain before pre-processing;(c) the frequency domain after preprocessing}
    \label{fig:preprocessing}
\end{figure}

The noise in the data acquired and stored by ADU-08e can be categorized into two distinct components. The first component is situated in the high-frequency band (greater than 30kHz) and emanates from the sampling frequency. To effectively eliminate this, the implementation of a low-pass filter is recommended. The second component pertains to the low-frequency band (less than 1000Hz). Illustrated in Fig.\ref{fig:preprocessing}(a) and (b), this lower frequency range's noise primarily comprises a DC component and a 50Hz power frequency, along with its harmonics.

To obtain relatively clean VLF signal data from the initial configuration of the magnetometer MFS-07e, we employ averaging to remove the DC component, while a low-pass filter is utilized to discard noise beyond the very low-frequency range. Furthermore, a notch filter is applied to effectively remove the 50Hz power frequency and its harmonic interferences. These processes collectively constitute the preprocessing steps employed on the original captured data. The result of preprocessing is shown in Fig.\ref{fig:preprocessing}(c).

\section{Experimentation and Validation}  
\subsection{Implementation Details}

\begin{table}[htb]
\caption{The hyperparameters used for training}
\centering
\begin{tabular}{cc}
\hline
\textbf{Hyperparameter} & \textbf{Value} \\ \hline
Batch Size              & 10             \\ \hline
Epoch                   & 12             \\ \hline
Loss Function           & CrossEntrophy  \\ \hline
Optimizer               & Adam           \\ \hline
Learning Rate           & 0.0001         \\ \hline
Momentum                & 0.5            \\ \hline
\end{tabular}
\label{tab:hyperparameters}
\end{table}

The essential hyperparameters outlined in this paper are presented in Table\ref{tab:hyperparameters}. In this scenario, the batch size of the model processes 10 data samples, each comprising 1000 data points. The parameter epoch is 12, indicating that the model undertakes 12 complete iterations of forward and backward calculations. Notably, the epoch of the proposed method is roughly less than half the value employed in CNN models.

The model adopts the Cross-Entropy function for its loss computation and employs the Adam optimization algorithm. Following preliminary testing, the Learning Rate is set at 0.01, and Momentum is set to 0.5. These values collectively wield influence over the model's convergence rate and operational efficiency.

\subsection{Comparison with State-of-the-art Methods}

\begin{table*}[htb]
\caption{The F1 score results of SOTA}
\begin{tabular}{clllllllllllll}
\hline
Types        & Res& FPN& -CG(\%) & CG-IR(\%) & NBR(\%) & +CG(\%) & MP(\%) & +PBP(\%) & -PBP(\%) & -NBE(\%) & +NBE(\%) & SW(\%) & AVG(\%) \\ \hline    
\multirow{3}{*}{CNN}         & no       & no  & 0.0           & 70.0          & 80.0          & 83.0          & 80.0          & 88.0          & 74.0          & 87.0          & 90.0          & 84.0          & 76.0          \\
                             & yes      & no  & 85.0          & 84.0          & \textbf{86.0} &95.0 & 93.0          & 94.0          & 85.0          & 93.0          & \textbf{95.0}     & 90.0          & 90.0          \\ 
                             & yes      & yes & 86.0          & 88.0          & 85.0          & \textbf{96.0}         & 92.0          & 94.0          & 87.0          & 94.0          & \textbf{95.0}         & 90.0          & 91.0          \\ \hline 
\multirow{3}{*}{Transformer} & no       & no  & 83.0          & 81.0          & 78.0          & 92.0          & 86.0          & 88.0          & 78.0          & 91.0          & 89.0          & 89.0          & 86.0          \\
                             & yes      & no  & 83.0          & 82.0          & 82.0          & 81.0          & 82.0          & 87.0          & 73.0          & 85.0          & 92.0          & 90.0          & 89.0          \\
                             & yes      & yse & \textbf{88.0} & \textbf{88.0} & 85.0          & \textbf{95.0} & \textbf{97.0} & \textbf{95.0} & \textbf{88.0} & \textbf{94.0} & 94.0          & \textbf{93.0} & \textbf{92.0}\\ \hline
\end{tabular}
\label{tab:results of SOTA}
\end{table*}

\begin{figure*}[htb]
    \centering
    \includegraphics[height=6cm,width=1\linewidth]{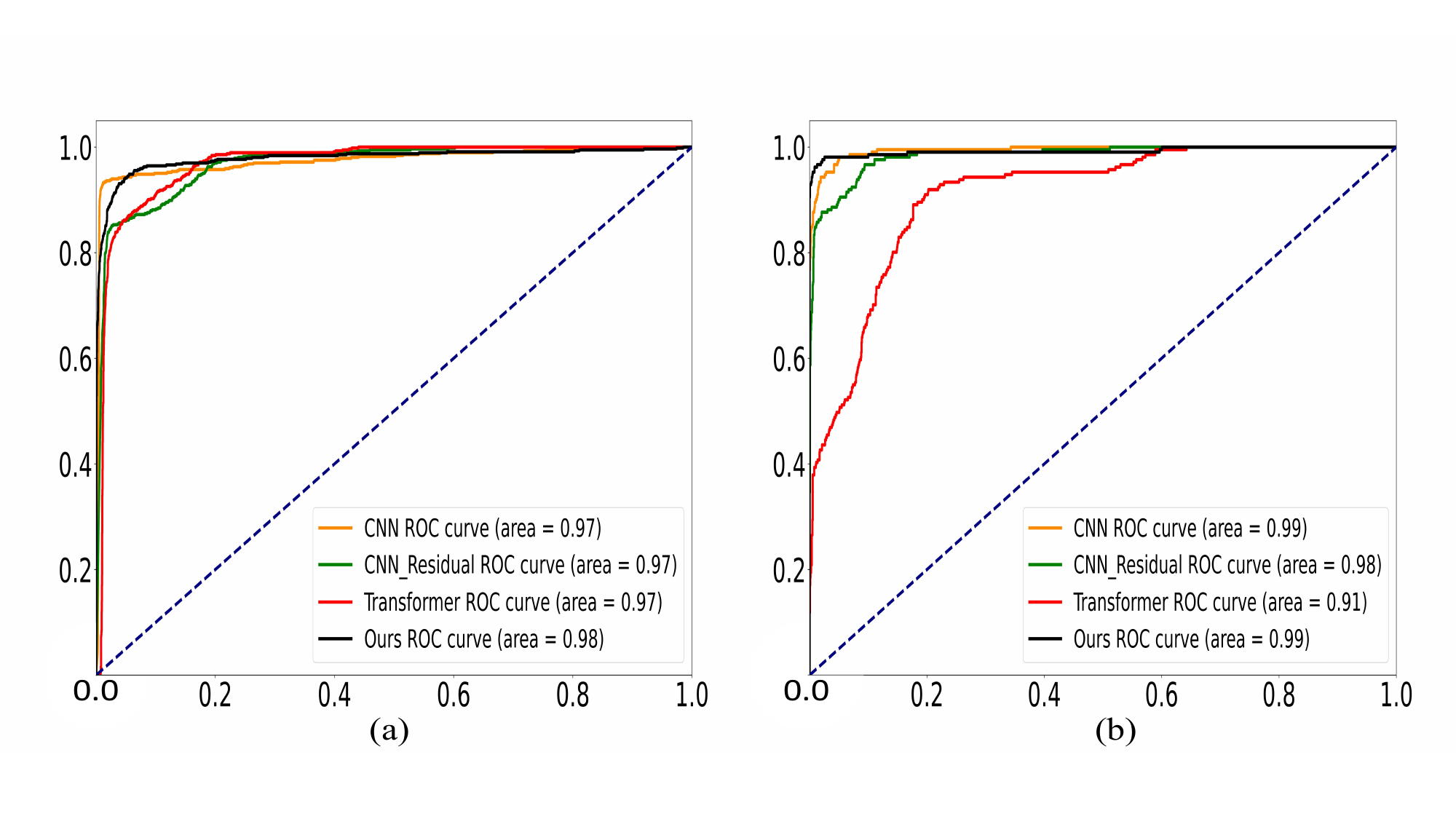}
    \caption{The comparison ROC for different lighting signals:(a) the ROC of CG lighting;(b)the ROC of -NBE lighting}
    \label{fig:ROC}
\end{figure*}

In this section, our method comparison with \textit{State-of-the-art} (SOTA) methods by using the experimental dataset and the F1-score as valuation criteria \ref{eq:F1socre}, which also known as the balanced score, is defined as the harmonic mean of precision and recall.
\begin{equation}
\label{eq:F1socre}
F1 = 2\cdot \frac{prcision \cdot recall}{prcision + recall} 
\end{equation}
Where the \textit{precision} represents the accuracy of predictions in the positive results of the sample. The \textit{recall} represents the proportion of true positive samples that are predicted correctly out of all actual positive samples in the sample set.

The function of the average F1 score is definite as:
\begin{equation}
\label{eq:Average F1}
Average F1 = \frac{\sum F1score}{Number\_ of\_ Epcho}
\end{equation}
Where $\sum F1score$ is the sum of the F1 score for each epoch, and the \textit{Number$\_$of$\_$Epoch} is the total number of epochs for training.

\subsubsection{Comparison with F1 score}

Table\ref{tab:results of SOTA} provides a comprehensive view of the average F1 scores achieved by state-of-the-art models for various types of lightning signals, all trained under the same epoch setting (epoch = 12). The models considered in the analysis encompass a spectrum of architectures, including the baseline CNN, CNN augmented with a residual module, CNN with both a residual and FPN module, Transformer, Transformer with a residual module, and our novel proposed method.

In both CNN and Transformer networks, the incorporation of residual connections and FPN modules improves the ability of signal classification. This observed improvement can be attributed to the intrinsic advantages of residual connections in addressing the vanishing gradient problem, promoting quicker convergence during training, and bolstering the model's ability to generalize effectively to previously unseen data. Simultaneously, the FPN module enriches the models with a mechanism for multi-scale feature fusion.

We compare our proposed method with other state-of-the-art methods, To ensure the fairness of the comparison, the datasets and training parameters remain consistent. In the majority of classification results, our method performs the superior performance. In "NBR" and "+NBE", our method's F1 score is close to the highest F1 score. These scores surpass the corresponding F1 scores obtained by other methods, highlighting the efficacy of our approach in significantly improving classification accuracy for these specific categories.

Aggregate metrics further corroborate the superiority of our proposed method. Across all considered evaluation criteria, our model achieves an outstanding average F1 score of 92$\%$. In contrast, competing models such as the baseline CNN only achieve 76$\%$, CNN with residual connections achieve 90$\%$, CNN with both residual and FPN modules achieve 91$\%$, the Transformer achieves 86$\%$, and Transformer with residual connections achieves 89$\%$. This consistent and substantial superiority across a diverse range of signal categories not only validates but further emphasizes the overall effectiveness of our proposed model in the challenging task of lightning signal classification.

\subsubsection{Comparison with Receiver Operating Characteristic}

To facilitate a comprehensive comparison of diverse model performances, the \textit{receiver operating characteristic} (ROC) methodology was employed. This approach utilizes a graphical plot, depicted in Fig. \ref{fig:ROC}, which showed the diagnostic prowess of a classifier system across varying discrimination thresholds.

In this study, the two specific lightning types: CG-IR in Fig. \ref{fig:ROC}(a)and -NBE in Fig. \ref{fig:ROC}(b) were selected. By comparing the accuracy of state-of-the-art (SOTA) models with ours using their respective ROC curves as a benchmark, our model demonstrated a higher accuracy with areas of 0.98 and 0.99 for the classification of the two lightning types.

In the context of the two aforementioned experiments, firstly, these experiments performed that our proposed model consistently outperforms its counterparts across a diverse spectrum of categories, thereby underscoring its comprehensive superiority in the realm of lightning signal classification. secondly, because of the effectiveness of the Transformer-related model in strategically evaluating the significance of various elements within the input lightning signal sequence. This inherent ability contributes to an overall classification accuracy that surpasses that of CNN and related models.

To further explore the potential performance improvements of the Transformer model, especially when integrated with the multi-scale component, we conducted an ablation experiment. The outcomes of this experiment are presented below.

\begin{figure*}[htb] 
    \centering
    \includegraphics[height=6cm,width=1\linewidth]{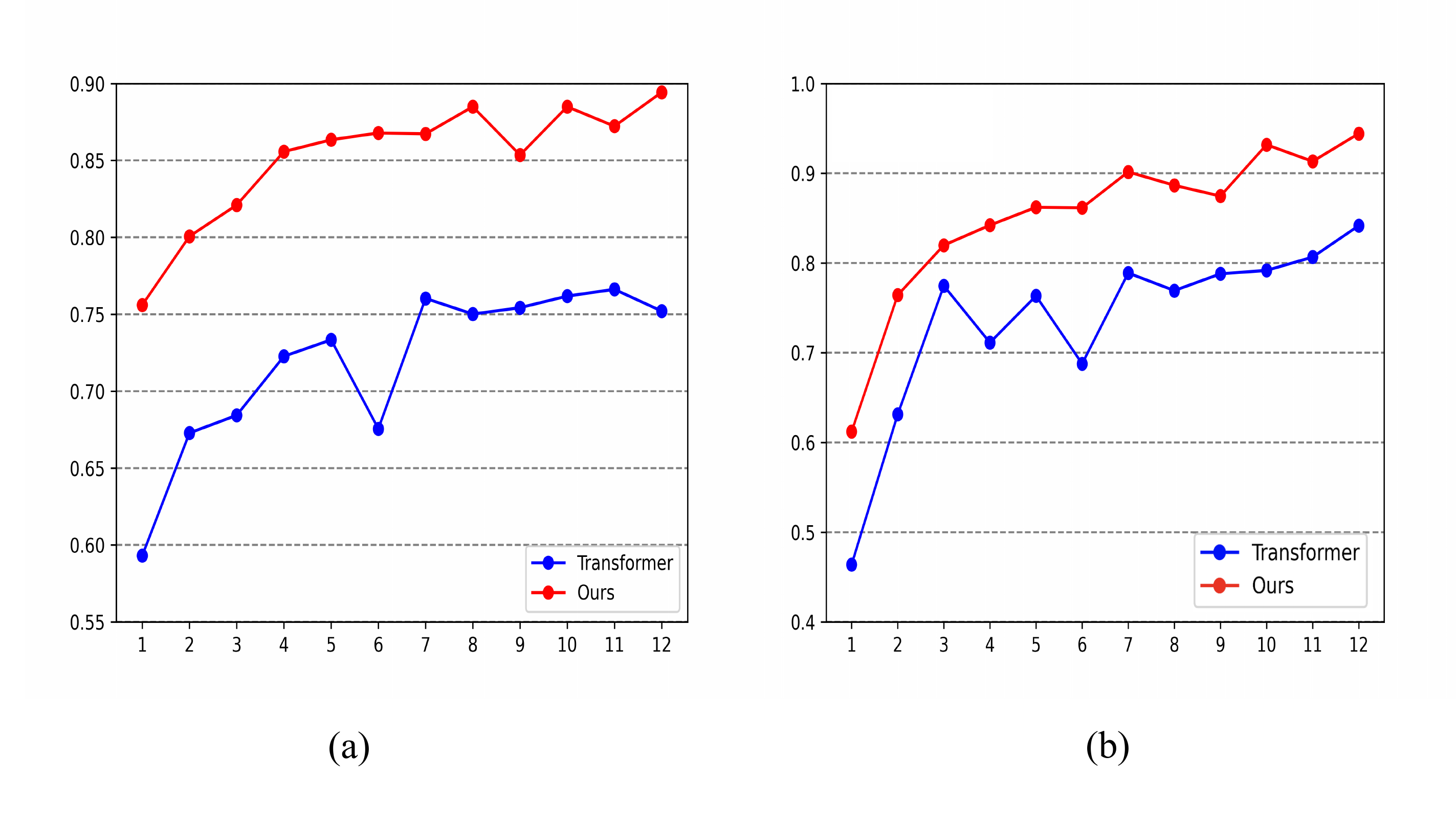}
    \caption{Comparison of transformer and our method,epoch=12:(a)the results of accuracy;(b)the results of F1 score}
    \label{fig:Comparision}
\end{figure*}

\begin{figure*}[htb] 
    \centering
    \includegraphics[height=8cm,width=1\linewidth]{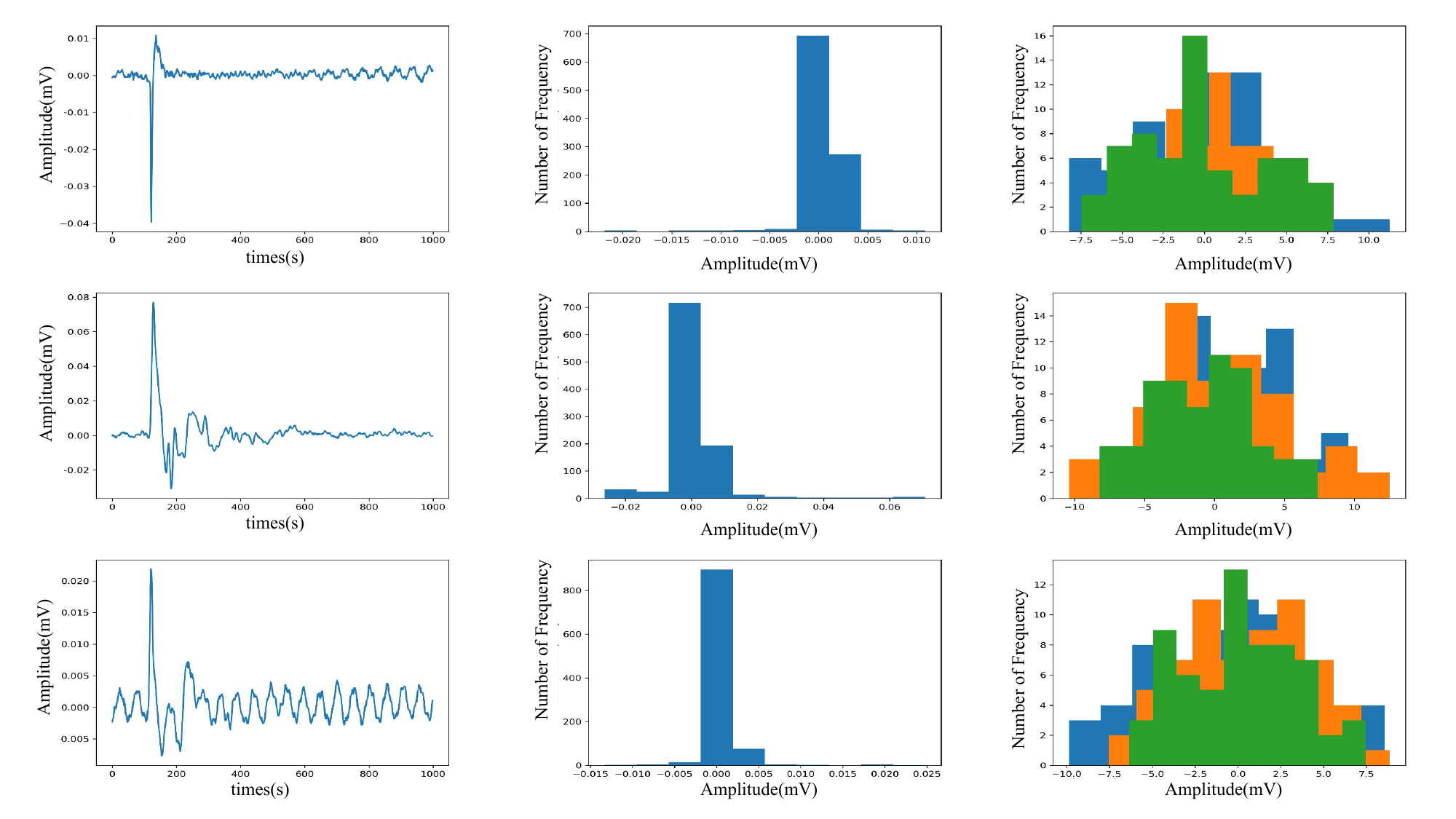}
    \caption{Different amplitude distributions of lighting signal at different scales: (Left Column)lightning signals; (Middle Column) histogram of lightning signals; (Right Column)histogram of signals at different scales}
    \label{fig:multi_fearturemap}
\end{figure*}

\subsection{Ablation Study}

In this section, we conducted a comprehensive analysis. Firstly, we compared with the Transformer model in terms of accuracy and F1 score. Subsequently, we examined the variations in feature maps within the MSR model at different scales. Finally, we evaluated the dataset’s performance through a five-fold cross-validation procedure, employing experimental datasets.

\subsubsection{Model Comparison}

Fig.\ref{fig:Comparision} presents a comparative analysis of accuracy and F1 scores attained by the Transformer model and our model across 12 epochs. Our method impressively achieves an accuracy of roughly 85$\%$ and F1 score of 90$\%$, markedly outperforming the Transformer model's accuracy of approximately 75$\%$ and F1 score 80$\%$. Evidently, our MSR model plays a pivotal role in the fusion of multi-scale lightning signal features throughout the classification process.

This discrepancy underscores the strength of our model in lightning signal classification. One of the advantages of this feature fusion process is its ability to capture detailed information that might otherwise remain hidden. Lightning signals often exhibit intricate variations across scales, and our model’s capacity to recognize and interpret these subtleties is contributing to the improvement in classification accuracy, ultimately leading to improvement in classification accuracy.

\subsubsection{Multi-scale Visualization Features Map}

Fig. \ref{fig:multi_fearturemap} provides a visual representation of distinct amplitude distributions within the lightning signal across a range of scales. This visualization is useful in understanding how the signal's characteristics evolve as we consider different scales.

The figure shows the gradual transition towards a more uniform amplitude profile as the scales expand. This phenomenon can be attributed to the influence of the multi-scale module within our model architecture. The module plays a pivotal role in harmonizing signal amplitudes and mitigating extreme amplitude disparities between individual lightning pulses.

The flattening of the amplitude serves as a compelling indicator of the efficacy of the multi-scale model. This transformation signifies the model's proficiency in effectively isolating and accentuating important signal features while reducing the significance of less relevant ones. This improvement in signal feature discrimination enhances the model's ability to distinguish between different features.

\subsubsection{Five-fold Cross-validation Results}

\begin{table*}[htb]
\caption{The F1 score of five-fold cross-validation for different types of classification}
\begin{tabular}{cccccccccccc}
\hline
K &  -CG(\%) & CG-IR(\%) & NBR(\%) & +CG(\%) & MP(\%) & +PBP(\%) & -PBP(\%) & -NBE(\%) & +NBE(\%) & SW(\%) & AVG(\%) \\ \hline
1 & 88      & 88        & 85      & 95      & 97     & 95       & 88       & 94       & 94       & 93     & 92      \\ \hline
2 & 87      & 87        & 86      & 96      & 94     & 94       & 88       & 94       & 95       & 92     & 91      \\ \hline
3 & 87      & 87        & 85      & 94      & 93     & 93       & 85       & 92       & 96       & 91     & 90      \\ \hline
4 & 85      & 86        & 85      & 94      & 92     & 92       & 84       & 92       & 95       & 90     & 90      \\ \hline
5 & 85      & 84        & 84      & 90      & 88     & 91       & 82       & 93       & 94       & 89     & 88 \\ \hline    
\end{tabular}
\label{tab:five-flod cross-validation}
\end{table*}

Table  \ref{tab:five-flod cross-validation} shows the F1 score of our model across an array of lighting types, each evaluated of individual folds of cross-validation. The F1 scores for most categories such as +CG, +PBP, -NBE, and +NBE consistently exceed 90$\%$, indicating the model's accuracy in classifying lightning signals. The "AVG($\%$)" column represents the average F1 score across all five folds for each classification category. This average score consistently demonstrates the model's strong overall performance, with an average F1 score exceeding 90$\%$.

Thus the table provides a coherent view of our model's performance across different types of lightning signal classification tasks using a rigorous five-fold cross-validation approach. The model consistently achieves high F1 scores, both at the fold level and in the overall average. That consistency proved our model's capacity to deliver dependable performance.

\subsubsection{The Results of Testing Dataset}

\begin{table}[htb]
\caption{The F1 score of five-fold cross-validation for testing dataset}
\centering
\begin{tabular}{lllllll}
\hline
K & CC(\%) & CG(\%) & MP(\%) & NBE(\%) & OTHERS & AVG(\%) \\ \hline
1 & 88     & 93     & 83     & 53      & 67         & 83  \\ \hline
2 & 83     & 94     & 88     & 41      & 33         & 80   \\ \hline
3 & 86     & 91     & 80     & 23      & 20         & 76   \\ \hline
4 & 82     & 89     & 70     & 31      & 40         & 74   \\ \hline
5 & 83     & 89     & 78     & 45      & 64         & 79    \\ \hline
\end{tabular}
\label{tab:cross-validation for validation dataset}
\end{table}

To show the robustness of our model's generalization capabilities,  the Xingjiang dataset was employed for testing. Utilizing a similar procedure as the experimental dataset, the testing dataset was divided into five categories, each containing approximately 360 instances. The obtained results, displayed in Table \ref{tab:cross-validation for validation dataset}, showcasing the F1 scores achieved by our model in each fold of cross-validation.

The table shows that despite the smaller scale of the Xinjiang testing dataset compared to the experimental dataset, our model consistently attains an accuracy rate of around 80$\%$. This accuracy rate demonstrates the model's exceptional ability to maintain classification stability and perform at a high level of accuracy when dealing with datasets of varying sizes and characteristics.

Our model demonstrates versatile adaptability, making it well-suited for real-world applications where lightning signal data can vary in terms of volume, distribution, and signal characteristics. Our model's capacity to generalize effectively across datasets further solidifies its credibility as a robust and reliable solution for lightning signal classification in diverse environments.

\section{Conclusion}
Accurate classification of lightning signals holds paramount importance within the realm of VLF lighting signal processing. The advancements in deep learning present a viable avenue for achieving precise lightning signal classification. Beyond the pursuit of classification accuracy, real-world applications necessitate the factors of network simplification and operational stability.

Within this paper, we introduce a multi-scale residual transformer model for VLF lightning waveform classification. This model incorporates a multi-scale residual module, merging low-level and high-level features. 

This fusion enhances the model's ability to identify detailed patterns and capture overall patterns in the signal. Additionally, the module's proficiency in accommodating features across multiple scales empowers the model to flexibly adapt to signal variations, thereby enhancing its robustness against changing signal characteristics. Integrating this module with a transformer encoder fortified by self-attention mechanisms empowers the model to effectively weigh the significance of different aspects within the input lightning signal sequence, enabling precise classification by focusing on pertinent information.

Through rigorous validation using both open-source and testing datasets, we demonstrate the superior stability and reliability of the proposed network in comparison to conventional CNNs. Furthermore, the prospects for the future application of our model hold the potential to significantly advance our comprehension of lightning signal localization and waveform characteristics. As a result, this paper not only contributes to accurate classification methodologies but also lays the groundwork for broader insights into lightning phenomena.

\section*{Funding}
This work was supported by the Equipment Pre-research Key Laboratory Fund Project(6142403190304),

\section*{Acknowledgment}
Thanks to the Institute of China Electronics Technology Group Corporation No.22 Research Institute for their data support. The author would also like to thank the reviewers for their helpful feedback, which significantly improved the manuscript.

\bibliographystyle{IEEEtran}


\begin{IEEEbiography}[{\includegraphics[width=1in,height=1.25in,clip,keepaspectratio]{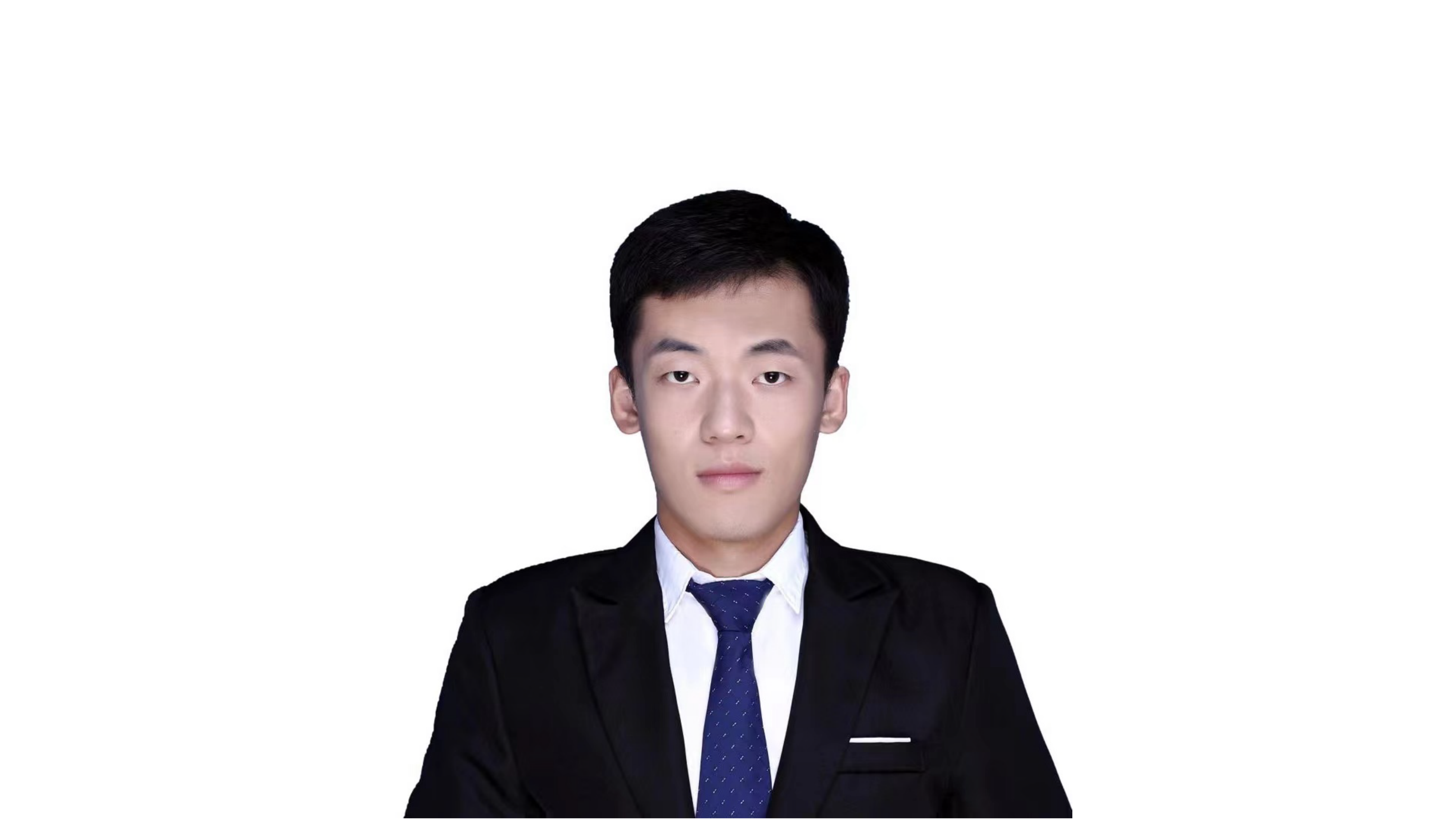}}]
{Jinghao Sun}
receive the B.Sc degree from Ocean University of China, Qingdao, China, in 2016. He is currently pursuing the Ph.D. degree in Faculty of Information Science and Engineering, Ocean University of China, Qingdao, China.
\end{IEEEbiography}
\begin{IEEEbiography}[{\includegraphics[width=1in,height=1.25in,clip,keepaspectratio]{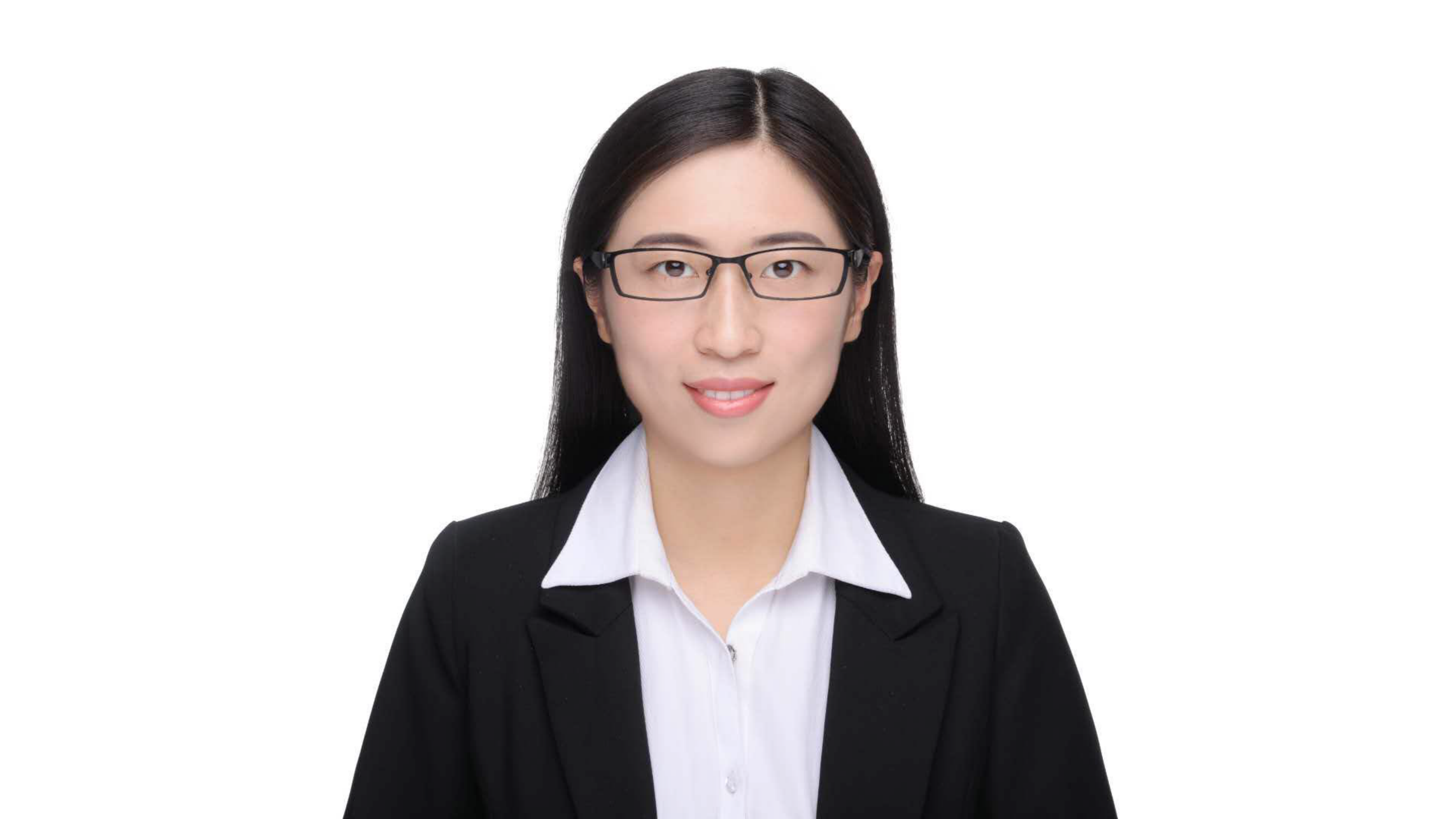}}]
{Tingting Ji}
received her PhD computer application technology from the Ocean University of
China, Qingdao, China, in 2010. She is now with the Teaching Center of Fundamental Courses,
Ocean University of China.
\end{IEEEbiography}
\begin{IEEEbiography}[{\includegraphics[width=1in,height=1.25in,clip,keepaspectratio]{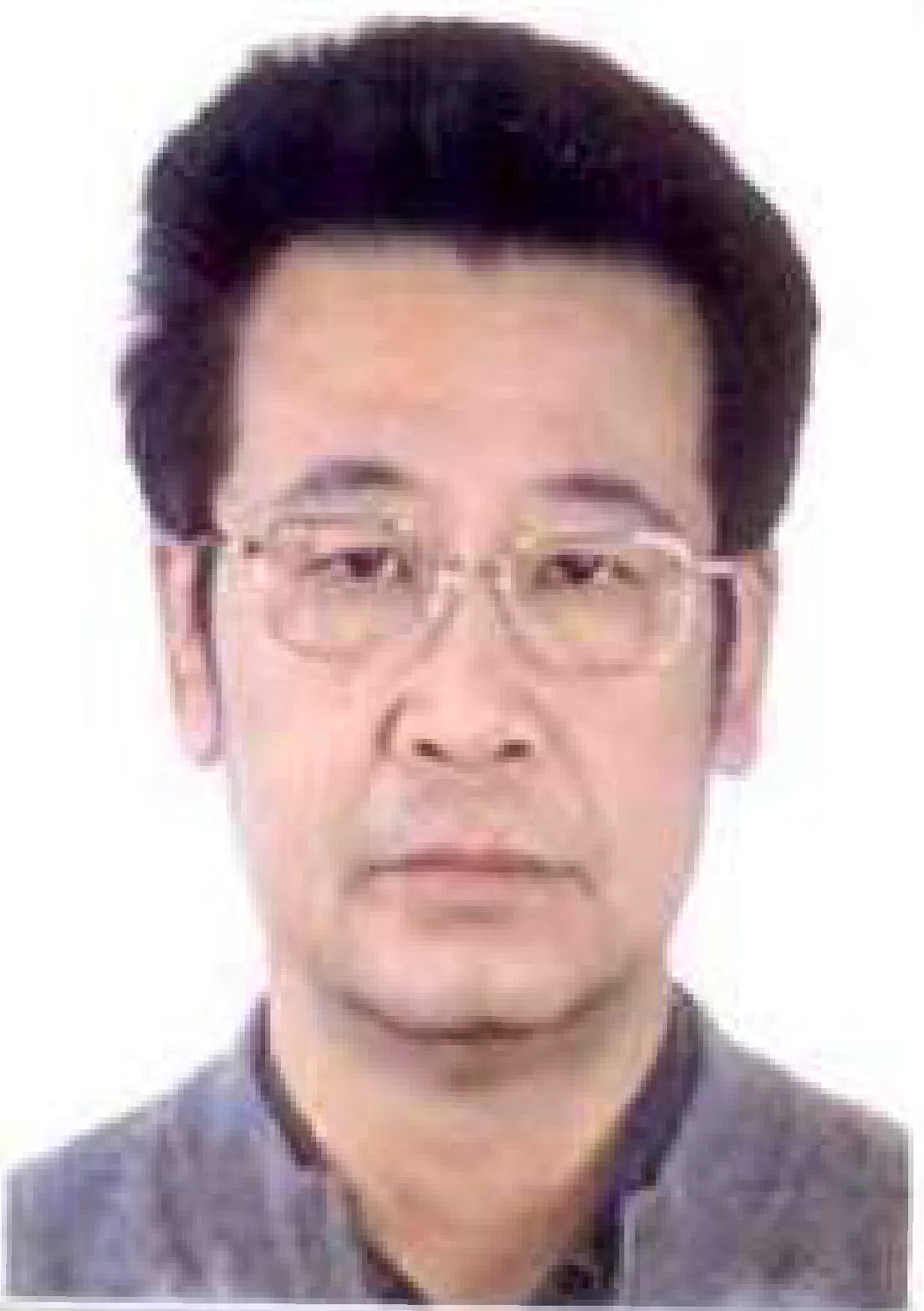}}]
{Guoyu Wang} 
received the B.S. and M.S. degrees in Ocean Physics from Ocean College of Shandong, Qingdao, China, in 1984 and 1987, respectively. He received Ph.D. degree in Faculty of Electrical Engineering from Faculty of Electrical Engineering, Netherlands, in 2000. He is now a Professor at Ocean University of China. His research interests include pattern recognition image processing and analysis.
\end{IEEEbiography}
\begin{IEEEbiography}[{\includegraphics[width=1in,height=1.25in,clip,keepaspectratio]{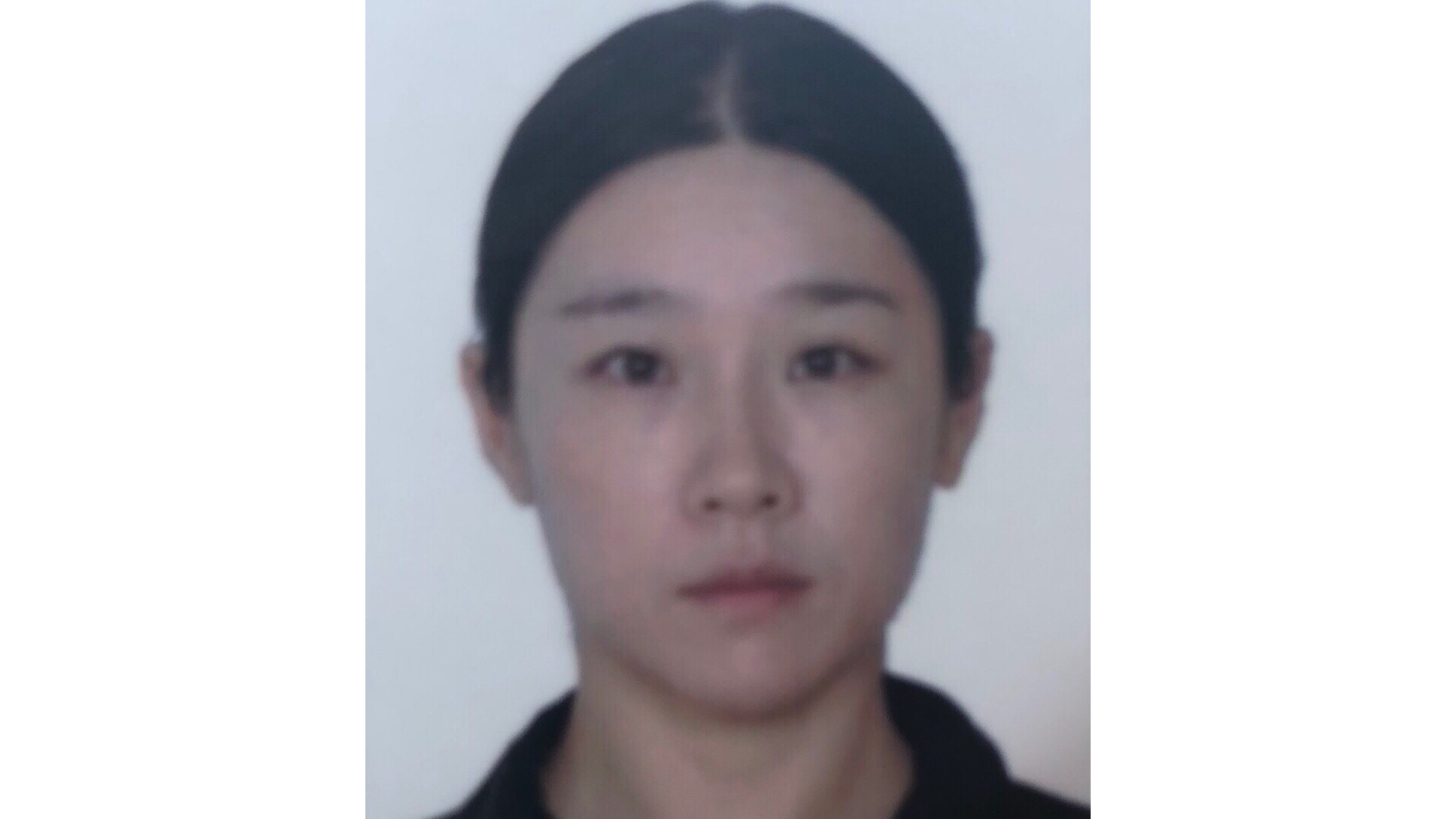}}]
{Rui Wang} 
received her BS degree in electronic information engineering and her PhD in computer application technology from the Ocean University of China, Qingdao, China, in 2009 and
2014, respectively. She is now at the School of Physics and Electronic Information of Yantai
University. Her research interests include image processing and computer vision.
\end{IEEEbiography}

\end{document}